# Knowledge Graphs and Machine Learning in biased C4I applications

Evangelos Paparidis, Konstantinos Kotis
Intelligent Systems Lab, Dept. of Cultural Technology and Communication
University of the Aegean
Mytilene, Greece
cti20009@ct.aegean.gr, kotis@aegean.gr

*Abstract*— **This paper introduces our position on the critical issue of bias that recently appeared in AI applications. Specifically, we discuss the combination of current technologies used in AI applications i.e., Machine Learning and Knowledge Graphs, and point to their involvement in (de)biased applications of the C4I domain. Although this is a wider problem that currently emerges from different application domains, bias appears more critical in C4I than in others due to its security-related nature. While proposing certain actions to be taken towards debiasing C4I applications, we acknowledge the immature aspect of this topic within the Knowledge Graph and Semantic Web communities**.

## I. Introduction

Nowadays, due to the large volume of data on the Web, there is a growing interest in knowledge graphs (KG), playing an important role in search engines e.g., Google. Early use of KGs has been conducted by Google in 2012 [1]. Using these graphs, simple word processing has become a symbolic representation of knowledge. KGs also are used by social networking and e-commerce sites and are of particular interest to the Semantic Web (SW) [2]. Although there is no accurate definition of a knowledge graph, it seems to better represent a database of interconnected descriptions of real-world entities and events or abstract concepts.

Although KGs are easily understood by humans and contain high-level information about the world, it is difficult to exploit them for Machine Learning (ML) applications. ML is one of the most significant fields of Artificial Intelligence (AI) research. Its goal is to create systems that can be trained from empirical data which had been observed in the past, to perform the work for which they are intended more effectively [3].

ML and KGs are developed in a fast pace. On the one hand, ML techniques improve the performance of various data-driven tasks with great precision. On the other hand, KGs provide the ability to represent knowledge about entities and their relationships with high reliability, explanation, and reuse. Consequently, a combination of KGs and ML will systematically improve systems' accuracy, explainability, and reuse, expanding the range of ML capabilities [4].

In the past decade, several research efforts had been conducted in the field of C4I (Command, Control, Communications, Computers & Intelligence) systems. C4I systems are of particular interest because they concern modern technological means of communications, information technology, and physical security which, with fully ensured interoperability, provide trustworthy information to authorized users to support real-time situational awareness, and facilitate data-driven knowledge-based decision-making.

Bias in research concerns the interference (as 'noise') in the outcomes of research (e.g., AI research) by predetermined ideas, prejudice or influence in a certain direction. Data in ML algorithms used in AI systems can be biased, but so can the algorithms that analyze this data. Both data (samples) and algorithms are designed by people, and people can be biased. When data is biased, the sample is not a perfect representative of the used dataset(s) in the algorithmic analysis. A recent case identified under the term of representational bias such as Google's image search for "CEO" depicting mostly males, may "teach" an intelligent system to recommend doctor as a career choice for men and nurse for women.

As there is no relevant research (to the best of our knowledge) that combines KGs and ML in C4I application domain, we have been motivated by the potential combination of those technologies to draw conclusions about their role in (de)biased AI applications. Our contributed conclusion on this topic, as presented in this paper, is that new advancements in knowledge and data representation are needed that will be used in C4I application domain to improve ML and data analysis in ways that critical problems such as data and algorithmic bias are reduced (or even eliminated).

This paper is structured as follows: Section II presents the topics of KGs, ontologies, and ML, Section III introduces the C4I application domain in respect to those technologies, open issues and research challenges are discussed in Section IV, and Section V concludes the paper.

## II. Knowledge graphs, ontologies, and machine learning

### A. Knowledge Graphs

There have been several attempts to define what a KG is. Due to the different definitions already present in the literature, some inconsistencies have been inevitably emerged. In addition to the definition in Wikipedia, other definitions have been proposed by various researchers [5]–[7]. A KG is the organization and representation of knowledge as a multi-domain graph, whose nodes represent entities of interest, combining different sources of controlled vocabularies and data.

Some of the most well-known non-public KGs are Google KG, Microsoft KG, and Facebook KG. There is also a large number of KGs available for the public such as DBpedia and Wikidata KGs. For the most KGs, there are no explicit references regarding the methods of extracting knowledge represented in those KGs, nor about their general shape, visualization, and storage of the available knowledge [8]. There are two approaches for creating a KG: a) top-down (schema-driven) and b) bottom-up (data-driven). In the first approach, the schema (ontology) of the graph are first defined and then the data is entered (schema population). In the second, the knowledge is extracted from various sources (e.g., text documents, DBs, as well as from linked open data) and after being merged, the schema (shape) of the KG is constructed.

Specifically, Facebook's KG is an essential tool enabling internal search within the Facebook platform to produce more and more accurate results when used by connected users. Google's KG is a valuable tool for Google Search, a knowledge base (KB) that gathers information from a variety of sources such as Wikipedia, in order to produce better and more complete results of the search engine. DBpedia's KG is a huge KB created by processing the information of Wikipedia. Wikipedia, because of the plethora of pages it contains, and especially those in different languages, there are contradictions, creating inaccuracies in the information. The problem of managing this information is solved by Wikidata through the KG, where all languages are integrated into one version of Wikidata so that information could be linked to multiple languages at the same time. It also allows the existence of conflicting information by providing a system that organizes everything properly by citing information sources [6].

*B. Knowledge Graphs and Ontologies*

An ontology is a formal and explicit specification of shared and agreed conceptualizations. A domain ontology is related to specific domain knowledge e.g., an ontology for a museum, an ontology for security, an ontology for surveillance. An ontology may define only the schema of the represented knowledge (classes, relations between them, class restriction axioms) or the schema and the actual data that are semantically described by the defined schema. In the second case, the ontology is a populated one i.e., an ontology with populated classes. In other words, an ontology is a knowledge base that stores knowledge about domain-specific entities, and those entities are classified as instances/individuals of its ontological classes.

KGs and populated ontologies are similar in a way. They both are related to the Resource Description Framework (RDF) for representing data. They both represent domain knowledge using semantic relations (links/edges) between entities (nodes). Knowledge about entities is represented by a triple of the form subject, predicate, and object (SPO) statements, where subject and object are mostly entities (or values), and the predicate is a relationship between them (or an attribute of an entity).

On the other hand, KGs and populated ontologies have some differences in respect to their aim. KGs often contain large volumes of factual information (facts about represented entities) with less formal semantics (class restriction axioms, definitions). On the other hand, an ontology defines the terminology of the domain and the semantic relations between terms, making knowledge available for machine processing, whereas data is not the main concern at design time. In addition, knowledge graphs can also represent knowledge about multiple domains and therefore may contain more than one ontology.

*C. Knowledge Graphs and Machine Learning*

ML is a way for the machine to achieve intelligence by learning from the data that is provided as input to a process of data analysis, thus evolving a machine that performs tasks into a smart machine. A ML method generally means a set of specific types of algorithms that are suitable for solving a particular type of computational problem. Such a method addresses any constraints that the problem brings along with it.

The most popular ML methods are supervised learning, semi-supervised learning, and reinforcement learning. Reinforcement learning is also applied in deep learning. Deep reinforcement learning applied when it is not possible to feasible training autonomous decision-making by other methods [9], [10]. The learning process consists of three stages: the acquisition of data, the processing/analysis of data so as to find possible generalizations or specializations, and the use of processing results to perform the objective work.

ML in KGs and ontologies are used to solve problems such as link and type prediction, ontology enrichment, and completion [3]. In particular, because abstract reasoning is not applicable, and at the same time, while an ontology is consistent, the information in it may be incorrect in relation to a reference domain, ML methods were developed for the Semantic Web (SW) in order to solve this issue.

Large-scale KGs are knowledge-based graphs storing real-world information in the form of relationships between entities. In an automated KG construction method, triples are automatically extracted from unstructured text using ML and natural language processing (NLP) techniques. In recent years, automated methods have been gaining more attention, because other methods either have several limitations or do not scale well due to their dependence on human experts (human bias) [4].

*D. Machine Learning and Deep Learning*

Over the last few years Deep Learning (DL) has been used in a great number of ML tasks, ranging from image scene classification and video analysis to natural speech, language understanding and recognition. The data used within these tasks are characterized by various data types such as images, voice signals, feature vectors and are typically represented in the Euclidean space. DL approaches successfully manage the aforementioned data types build upon variant deep network architectures include but are not limited to Convolutional Neural Networks (CNNs) [11], Recurrent Neural Networks (RNNs) [12], Long Short-Term Memory (LSTM) [13] and auto-encoders [14]. Opposite to the architectures mentioned, there are a great number of DL-based architectures that the data used are projected to a non-Euclidean space, especially in the form of KGs [15]. This data structure overcomes the restrictions of interaction assumption following a linking approach. Thus, during the linking between the items and their attributes, each node is linked with various nodes and variant types.

Connecting nodes in a KG may have distinct neighborhood size, and the relationships between them could vary as well. The need to handle the above complexity of KGs stimulates new neural network architectures mentioned as Graph Neural Networks (GNNs) [16]. GNNs learn a target node's representation based on propagating the information to one or many neighbors in a recurrent way until a stable fixed point is reached [17]–[19]. This process lacks computational efficiency; thus, recently, there have been increasing efforts to overcome this limitation [20], [21]. These studies belong to the category of Recurrent Graph Neural Networks (RecGNNs).

### III. C4I APPLICATION DOMAIN

The Command, Control, Communications, Computer and Intelligence (C4I) systems are modern technological means of communications, information technology (IT) and physical security, which provide information to authorized security commanders, in order to be aware of the situation in real-time, supporting decision making. These systems do not replace the human factor or conduct security operations on their own. However, they are an important tool for providing timely and reliable information and facilitate the work of managers in decision making. According to the latest developments, C4I systems are based on static architectures, and thus they require dynamic features to provide the latest developments in IT, to ensure reliability, real-time support, ease of reuse and interoperability [22].

Standard Information Exchange Data Models (IEDMs) involve difficult mapping and model translation tasks that cannot be easily automated. Also, the multiple use of IEDMs creates the need for maintenance and expansion. Because manual extensions to data models lead to unsustainable extension management processes, there is an increased need for distributed management and scalability. A framework for such a model management methodology can be supported using OWL ontologies [23]. OWL ontologies have been used in the C4I domain already. An example of the usefulness of OWL-DL is the Cyber forensics ontology [24]. This ontology represents knowledge related to the extraction of evidences as well as the law of cybercrime. The aim of this ontology is to classify types of crimes in order to draw conclusions about cybercrime and to identify similar methods of criminal investigation.

Another area of particular interest is the Operational Headquarters (HQ). Officers are required to be aware of all the information around their reference object and at the same time maintain command and control (C2) in an increasing variety of operations. The need for interoperability between HQ information systems, so that information can be absorbed effectively by the government agencies to which they belong and simultaneously evaluated and presented in an integrated manner, led to the creation of the 5th Generation Operational Level Military Headquarters [25]. To solve the problem that arose, a microservice API was designed as an extension of the Saki architecture [25]. The embedded information is forwarded to a knowledge graph for long-term storage. However, over time, changes will need to be made to the way information in the KGs is modeled. This issue has been solved by committing to an ontology all possible areas of operations, and at the same time, all the components of the microsystem will exchange information using this ontology [25].

The Semantic Web technology targets to make the Web readable from the machines [26], by enriching resources with metadata. To manage these metadata OWL models are used, while the reasoning capabilities provided by the ontologies are also taken into account. However, metadata management process meets significant limitations -especially in cases of linked data- that include but are not limited to the time consumption for ontology construction, inconsistent and noisy data.

The problems of query answering, instance retrieval, link prediction, and ontology refinement and enrichment have also emerged, with the three firsts considered by ML methods as classification problems while the last as a concept learning problem. ML methods' introduction to solving classification and concept learning problems on the Semantic Web domain considers the advantages of the numeric-based (sub-symbolic) approaches such as DL [27] and embeddings. These ML methods are generally categorized into symbolic and numeric-based. ICAS cyber-security ontology utilizes a tool called Targeted Attack Premonition using Integrated Operational data (TAPIO) which was implemented to export in real time all the data of each company automatically into a fully linked semantic graph [28]. This tool uses ML techniques, among other domains, to implement name entity resolution. Also, in order to facilitate the storage of huge data export and the visualization of important standards in C4I systems, the integration of large analytical data utilizing various tools that use ML to classify large data has been proposed [29].

ML and DL are widely used today in AI-based systems and applications to improve battle results and mission execution [30]. Such technologies are the key to the development of the Fleet's Command, Control, Communications, Computers, Information, Surveillance and Reconnaissance (C4ISR) architecture and processes.

### IV. OPEN ISSUES AND CHALLENGES

Today, in addition to the difficulties we experience every day, the rapid increase in unemployment and crime and social fluctuations, a very important and invisible problem has been added, i.e., the problem of bias in AI applications and systems. When input data is biased, the sample is not representative of the utilized dataset(s) in the algorithmic analysis.

Discussions within the ML community about how to address representational biases have not yet reached the KG and SW communities. The current status of Linked Open Data (LOD) cloud may be considered as free of selection biases such as sampling bias. However, the data available to both open and commercial KGs today is, in fact, highly biased. Debiasing KGs (data and schema) will soon become a major issue as these are now rapidly used in several ML-based algorithms of AI systems and applications.

Debiasing KGs must be examined at the level of data (data bias) as well as the level of schema (schema/ontology bias). Entities represented in DBpedia's KG, for instance, either spatial or non-spatial, do not cover the global range of available data (all the world as we know it), instead, the coverage of data related to Europe- and US-based entities is clearly larger than Asia's. Bias at the schema/ontology level, on the other hand, is also present, since most of the ontologies are engineered following a top-down methodological approach, often with application needs in mind. In such cases, a group of knowledge engineers/workers and domain experts collaborate, 'passing' their human biases in their engineering choices and the designing patterns (human-centric). Finally, if bottom-up (data-driven) ontological engineering approaches are used, for instance, based on ML algorithms that derive axioms/rules from data, the bias problem remains (data biases) as discussed before.

In this article we contribute our position regarding the need to propose a new methodology for the engineering of bias-free KGs, supported by suitable tools for managing KGs both at the level of data and schema, aligned with modern policies/rules on AI bias elimination. Based on such a methodology (distinct phases, processes and tasks), it is argued and expected that bias-free AI applications will prevail. In the C4I application domain, this is largely translated to algorithms that, for instance, classify human entities as terrorists/robbers/rapists based on features learned from non-biased data, and certainly not because they "accidently" have a black/dark colored hooded face and a short-height body that carries a particular type of a backpack.

## V. CONCLUSION

The magnitude of the problem of biased KGs is not fully realized today because we do not fully understand how it is affecting our daily lives. It is expected that in the next few years, it will seriously affect the lives of thousands worldwide. From a mis-rejected job application to the false arrest of an innocent fellow and the mis-identification of the actual criminal or threat, debiasing AI applications must be a priority and a continuous concern of actions to be taken in the era of a globalized and secure world. Biased KGs must be treated now, at early stages, to avoid fast spread over time. The society should not reach the extreme point where there is an emergency to clean biased data and schemas of the cult of public security.

In this position paper, the problem of bias in KGs and ML algorithms in the C4I domain has been introduced, concluding to actions that must be taken today by the related research and development communities. Towards these actions, our future work will be focusing on the design of a tool-supported methodology for the engineering of bias-free KGs.

REFERENCES

[1] H. Paulheim, "Knowledge graph refinement: A survey of approaches and evaluation methods," IOS Press, 2017. doi: 10.3233/SW-160218.
[2] P. Bonatti, S. Decker, A. Polleres, and V. Presutti, "Knowledge Graphs: New Directions for Knowledge Representation on the Semantic Web (Dagstuhl Seminar 18371)," *Dagstuhl Reports*, vol. 8, no. 9, pp. 29–111, 2019, doi: 10.4230/DagRep.8.9.29.
[3] C. D'Amato, "Machine Learning for the Semantic Web: Lessons learnt and next research directions," 2020. doi: 10.3233/SW-200388.
[4] M. Nickel, K. Murphy, V. Tresp, and E. Gabrilovich, "A review of relational machine learning for knowledge graphs," 2016. doi: 10.1109/JPROC.2015.2483592.
[5] L. Ehrlinger and W. Wöß, "Towards a definition of knowledge graphs," *CEUR Workshop Proceedings*, 2016.
[6] M. Färber, F. Bartscherer, C. Menne, and A. Rettinger, "Linked data quality of DBpedia, Freebase, OpenCyc, Wikidata, and YAGO," *Semant. Web*, vol. 9, no. 1, pp. 77–129, 2018, doi: 10.3233/SW-170275.
[7] A. Blumauer, "From Taxonomies over Ontologies to Knowledge Graphs - Semantic Web Company," *Semantic Web Company*, 2014.
[8] Z. Zhao, S.-K. Han, and I.-M. So, "Architecture of Knowledge Graph Construction Techniques," *Int. J. Pure Appl. Math.*, vol. 118, no. 19, pp. 1869–1883, 2018.
[9] V. Mnih *et al.*, "Human-level control through deep reinforcement learning," *Nature*, vol. 518, no. 7540, pp. 529–533, 2015, doi: 10.1038/nature14236.
[10] J. X. Chen, "The Evolution of Computing: AlphaGo," *Computing in Science and Engineering*, vol. 18, no. 4. IEEE Computer Society, pp. 4–7, Jul. 01, 2016, doi: 10.1109/MCSE.2016.74.
[11] Y. Bengio, "Convolutional Networks for Images, Speech, and Time-Series Oracle Performance for Visual Captioning View project MoDeep View project," 1997.
[12] A. L. Caterini and D. E. Chang, "Recurrent neural networks," *SpringerBriefs Comput. Sci.*, vol. 0, no. 9783319753034, pp. 59–79, 2018, doi: 10.1007/978-3-319-75304-1_5.
[13] S. Hochreiter and J. Schmidhuber, "Long Short-Term Memory," *Neural Comput.*, vol. 9, no. 8, pp. 1735–1780, Nov. 1997, doi: 10.1162/neco.1997.9.8.1735.
[14] P. Vincent, H. Larochelle, I. Lajoie, Y. Bengio, and P. A. Manzagol, "Stacked denoising autoencoders: Learning Useful Representations in a Deep Network with a Local Denoising Criterion," 2010.
[15] Y. Gao, Y. F. Li, Y. Lin, H. Gao, and L. Khan, "Deep learning on knowledge graph for recommender system: A survey," 2020, *arXiv*: 2004.00387. [Online]. Available: http://arxiv.org/abs/2004.00387.
[16] Z. Wu, S. Pan, F. Chen, G. Long, C. Zhang, and P. S. Yu, "A Comprehensive Survey on Graph Neural Networks," *IEEE Trans. Neural Networks Learn. Syst.*, vol. 32, no. 1, pp. 4–24, 2021, doi: 10.1109/TNNLS.2020.2978386.
[17] M. Gori, G. Monfardini, and F. Scarselli, "A new model for earning in raph domains," *Proc. Int. Jt. Conf. Neural Networks*, vol. 2, pp. 729–734, 2005, doi: 10.1109/IJCNN.2005.1555942.
[18] F. Scarselli, M. Gori, A. C. Tsoi, M. Hagenbuchner, and G. Monfardini, "The graph neural network model," *IEEE Trans. Neural Networks*, vol. 20, no. 1, pp. 61–80, 2009, doi: 10.1109/TNN.2008.2005605.
[19] C. Gallicchio and A. Micheli, "Graph echo state networks," *Proc. Int. Jt. Conf. Neural Networks*, 2010, doi: 10.1109/IJCNN.2010.5596796.
[20] Y. Li, R. Zemel, M. Brockschmidt, and D. Tarlow, "Gated graph sequence neural networks," 2016.
[21] H. Dai, Z. L. Kozareva, B. Dai, A. J. Smola, and L. Song, "Learning steady-states of iterative algorithms over graphs," 2018.
[22] A. M. Kulkarni, V. S. Nayak, and P. V. R. R. B. Rao, "Comparative study of middleware for C4I systems Web Services vis- Vis Data Distribution Service," in *Proceedings of the 2012 International Conference on Recent Advances in Computing and Software Systems, RACSS*, 2012, pp. 305–310, doi: 10.1109/RACSS.2012.6212685.
[23] K. Gupton, J. Abbott, C. Blais, S. Y. Diallo, K. Heffner, and C. Turnitsa, "Management of C4I and M&S data standards with modular OWL ontologies," *Spring Simul. Interoperability Work.*, 2011, pp. 404–417.
[24] H. Park, S. H. Cho, and H. C. Kwon, "Cyber forensics ontology for cyber-criminal investigation," in *Lecture Notes of the Institute for Computer Sciences, Social-Informatics and Telecommunications Engineering*, 2009, vol. 8 LNICST, pp. 160–165, doi: 10.1007/978-3-642-02312-5_18.
[25] A. Zschorn, H.-W. Kwok, and W. Mayer, "Microservice API design to support C2 semantic integration," 2019.
[26] C. D'Amato, N. Fanizzi, and F. Esposito, "Inductive learning for the Semantic Web: What does it buy?," *Semant. Web*, vol. 1, no. 1–2, pp. 53–59, 2010, doi: 10.3233/SW-2010-0007.
[27] L. Deng and D. Yu, "Deep learning: Methods and applications," *Foundations and Trends in Signal Processing*, vol. 7, no. 3–4. Now Publishers Inc, pp. 197–387, 2013, doi: 10.1561/2000000039.
[28] M. Ben Salem and C. Wacek, "Enabling new technologies for cyber security defense with the ICAS cyber security ontology," 2015.
[29] V. Shukla, B. Singh, M. Kumar, and K. Negi, "Big Data Analytics in C4I Systems," *2018 Int. Conf. Autom. Comput. Eng. ICACE 2018*, pp. 102–106, 2018, doi: 10.1109/ICACE.2018.8687057.
[30] K. Rothenhaus, K. De Soto, E. Nguyen, and J. Millard, "Applying a DEVelopment OPerationS (DevOps) Reference Architecture to Accelerate Delivery of Emerging Technologies in Data Analytics, Deep Learning, and Artificial Intelligence to the Afloat U.S. Navy," 2018.